\documentclass[11pt]{article}
\usepackage{acl}
\usepackage{times}
\usepackage{latexsym}
\usepackage[T1]{fontenc}
\usepackage[utf8]{inputenc}
\usepackage{microtype}
\usepackage{inconsolata}
\usepackage{graphicx}
\usepackage{amsmath}
\usepackage{enumitem}
\usepackage{booktabs}
\usepackage{xcolor}
\usepackage{array}
\usepackage{multirow}
\usepackage{makecell}
\usepackage{mdframed}
\usepackage{natbib}
\definecolor{admitgreen}{RGB}{22, 101, 52}
\definecolor{rejectred}{RGB}{159, 18, 57}
\definecolor{lightgray}{RGB}{248, 249, 250}
\definecolor{quoteblue}{RGB}{59, 130, 246}

\newmdenv[
  leftline=true, rightline=false, topline=false, bottomline=false,
  linewidth=3pt, linecolor=quoteblue,
  backgroundcolor=lightgray,
  innerleftmargin=8pt, innerrightmargin=6pt,
  innertopmargin=4pt, innerbottommargin=4pt,
  skipabove=4pt, skipbelow=4pt
]{examplebox}

\newcommand{\admit}[1]{\textcolor{admitgreen}{\textbf{[ADMIT]}} \textit{#1}}
\newcommand{\reject}[1]{\textcolor{rejectred}{\textbf{[REJECT]}} \textit{#1}}

\title{Gated Memory: Admission-Controlled Memory Formation for Conversational AI}

\author{
  Preeti Saraswat$^{\dagger}$ \quad Divya Neelagiri$^{*}$ \quad Ajay Manoj$^{*}$ \\
  Samsung Research America, Mountain View, CA \\
  \texttt{\{p.saraswat, d.neelagiri, ajay.manoj\}@samsung.com}
}

\begin{document}
\maketitle

\begin{abstract}
Personalized conversational AI relies on long-term memory systems that extract
facts from user utterances and stores them in persistent vector stores.
Despite substantial progress in retrieval, deduplication, and lifecycle
management, the \emph{formation stage},the moment a fact is first written to
storage has received almost no principled attention.
We identify this as the binding constraint on memory 
quality in production deployed systems. Critical contextual signals 
such as the distinction between a permanent user attribute 
and a transient situation, exist only in the original utterance 
and are irreversibly lost the moment extraction produces a 
subject-relation-object triple. No downstream process can 
recover them.
We propose \textbf{Gated Memory}, a lightweight, modular formation framework
that interposes two decision checkpoints between conversation and storage:
an \emph{admission gate} that evaluates every candidate fact against
the full utterance context before extraction runs, and a \emph{conditional
enrichment stage} that grounds admitted facts through an entity scope taxonomy
with privacy constraints. The gate evaluates only the current exchange while
using prior turns as read-only reference context, and produces a structured
formation record: admitted content is decomposed into atomic facts, each
categorized, tagged with provenance (directly stated versus inferred), scoped to
its condition of applicability, and grounded in resolved time and place, subject
to a constraint that no entity absent from the context may be asserted.
On the LoCoMo-10 long-term memory benchmark with atypical emotional density in utterance data, Gated Memory achieves an overall
\textbf{+2.6\% relative improvement} in LLM-judge accuracy over a strong
baseline with identical retrieval and generation, establishing formation quality as a measurable constraint 
with direct implications for actual production memory systems that might be polluted with a lot of irrelevant context.
\end{abstract}

% \noindent\textbf{Keywords:}
% long-term memory, conversational AI, long context, admission control,
% retrieval-augmented generation, personalization

% %% ────────────────────────────────────────────────────────────────
\section{Introduction}
\label{sec:intro}

Long-term memory in conversational AI is fundamentally a write problem.
The decisions made at the moment of storage,what to commit and in what
form,determine the ceiling of every subsequent interaction.
Systems such as MemGPT~\citet{memgpt}, mem0~\citet{mem0}, and
A-MEM~\citet{amem} have invested heavily in post-storage operations:
retrieval-augmented generation~\citet{lewis2020rag}, deduplication, lifecycle
management, and episodic/semantic categorization~\citet{tulving1985}.
These contributions are genuine, but they share an unexamined assumption:
that facts entering the store are worth storing.

This assumption fails in a structurally interesting way. When a user says
\textit{``I drove an SUV last weekend ,the only rental they had,''} any
extraction model produces \texttt{(User, drives, SUV)} ,semantically
identical to the triple for a vehicle the user actually owns. The signal that
separates them, the word \textit{rental}, exists once in the original
utterance. Once extraction discards the utterance and retains only the triple,
the ownership signal is permanently gone. Better retrieval cannot distinguish
two identical triples,better generation cannot reason correctly over
contradictory ground truth,pruning can only remove a stale entry long after
it has shaped responses across multiple sessions.

\textbf{This is a formation problem.} The evidence for the correct decision
ceases to exist downstream. The right intervention is at the only moment it is
possible: before extraction, while the full utterance is intact.

We introduce \textbf{Gated Memory}, a modular, admission-controlled formation layer
interposed between conversation and storage. Our contributions are:

\begin{itemize}[leftmargin=*,itemsep=1pt,topsep=3pt]
  \item We identify \emph{formation} as a structurally distinct, previously
  unaddressed failure point in long-term memory, characterizing three failure
  modes: reactive admission, absent enrichment, and irrecoverable loss of context.
  \item We propose an \emph{utterance-level admission gate} that evaluates
  candidate facts against the full original utterance before extraction ,the
  only viable point of intervention. The gate operates over a
  \emph{current-versus-reference} decomposition of the dialogue: only the current
  exchange is a source of new facts, while prior turns serve as read-only context
  for reference resolution, preventing the re-emission of previously stored facts.
  \item We formalize formation as \emph{structured extraction rather than
  binary filtering}: admitted content is decomposed into one or more atomic facts
  (multi-intent splitting), each assigned one or more of six formation categories,
  a provenance label distinguishing directly stated from inferred facts, and a
  conditional scope recording when the fact applies.
  \item We introduce a four-class \emph{entity scope taxonomy}
  (\texttt{USER\_PII}, \texttt{INTERNAL}, \texttt{EXTERNAL}, \texttt{GENERIC})
  governing conditional enrichment with privacy guarantees, together with
  \emph{deictic grounding} of relative temporal and spatial references at
  formation time.
  \item We enforce a \emph{grounding constraint} that forbids the assertion of any
  entity or relation not present in the evaluated window, eliminating a class of
  fabrication errors that extraction otherwise introduces, and we preserve the
  \emph{language} of the user utterance in the stored fact.
  \item We demonstrate a +2.6\% relative improvement in LLM-judge accuracy
  on LoCoMo-10~\citet{locomo} with zero changes to retrieval, generation, or
  evaluation, isolating the contribution of formation quality.
\end{itemize}

%% ────────────────────────────────────────────────────────────────
\section{The Memory Formation Problem}
\label{sec:problem}

\paragraph{Formation errors are undetectable downstream.}
All major memory frameworks share a common pipeline: extract
subject-relation-object triples, resolve conflicts via lifecycle reasoning
(ADD/UPDATE/DELETE/NOOP), and commit to a vector store. The implicit contract
is that extraction errors will be caught by downstream deduplication or
pruning. This contract fails in two ways. First, many formation errors are
\emph{undetectable}, a rental vehicle and an owned vehicle produce
non-conflicting, non-duplicate triples there is nothing for deduplication to
flag. Second, detectable errors persist across multiple sessions before pruning
fires, actively shaping system responses in the interim. Beyond transient
facts, semantically vacuous content pollutes the retrieval pool. The following
entries were produced by an existing framework on LoCoMo:

\begin{examplebox}
\textit{``Life has been ok, taking care of things''} \\
\textit{``Has favorite memories''} \\
\textit{``Appreciates little moments''} \\
\textit{``Has been busy with the project''}
\end{examplebox}

These entries pass deduplication (they are unique), survive pruning (they carry
no expiration signal), and occupy top-$k$ retrieval slots on virtually every
life-related query. The system is not malfunctioning ,it simply has no gate.

\paragraph{Admission must precede extraction.}
Modern systems address formation quality through lifecycle reasoning: an LLM
evaluates each extracted fact against the existing store and decides whether to
add, update, delete, or ignore it as existing fact. This is valuable for store consistency, but
it solves a different problem. Lifecycle reasoning asks whether a fact
\emph{conflicts} with stored knowledge,admission control asks whether a fact
\emph{deserves to be stored at all} ,whether it is temporally stable,
user-specific, and reusable. More critically, lifecycle reasoning operates on
the \emph{extracted triple}, not the original utterance. By the time it runs,
the context is already gone. \textbf{Admission must precede
extraction.} This is the architectural insight that all existing systems miss.

The same argument applies beyond the admit/reject decision. Whether a fact is
directly asserted or merely implied in passing, whether an utterance carries one
storable proposition or several, and which antecedent an unbound reference should
resolve to, are all properties of the original utterance that are erased the
moment extraction produces a triple. A system that reasons only over triples
cannot recover them, and a system that reconsumes the entire dialogue history on
every turn compounds the problem: it re-emits facts it has already stored and
resolves references against stale antecedents, fabricating entities the current
turn never introduced. Correct formation therefore requires not only admitting
the right facts, but categorizing them, attributing their provenance, and scoping
their evaluation to the current exchange,all before extraction discards the
evidence.

\paragraph{The enrichment gap.}
Even well-admitted facts are stored as raw extraction strings with no semantic
grounding. \textit{``User drives Car-X''} cannot answer queries about
fuel infrastructure or type of car. Enrichment done once at formation for stable facts is
amortized across every future retrieval,enrichment deferred permanently leaves
facts shallower than they need to be. This problem also interacts with privacy.
Uniform enrichment risks sending personal information to external
services without consent, while zero enrichment forfeits semantic depth. The
correct solution is conditional enrichment governed by scope classification at
formation time.

%% ────────────────────────────────────────────────────────────────
\section{Gated Memory}
\label{sec:method}

\begin{figure*}[t]
  \centering
  \includegraphics[width=\textwidth]{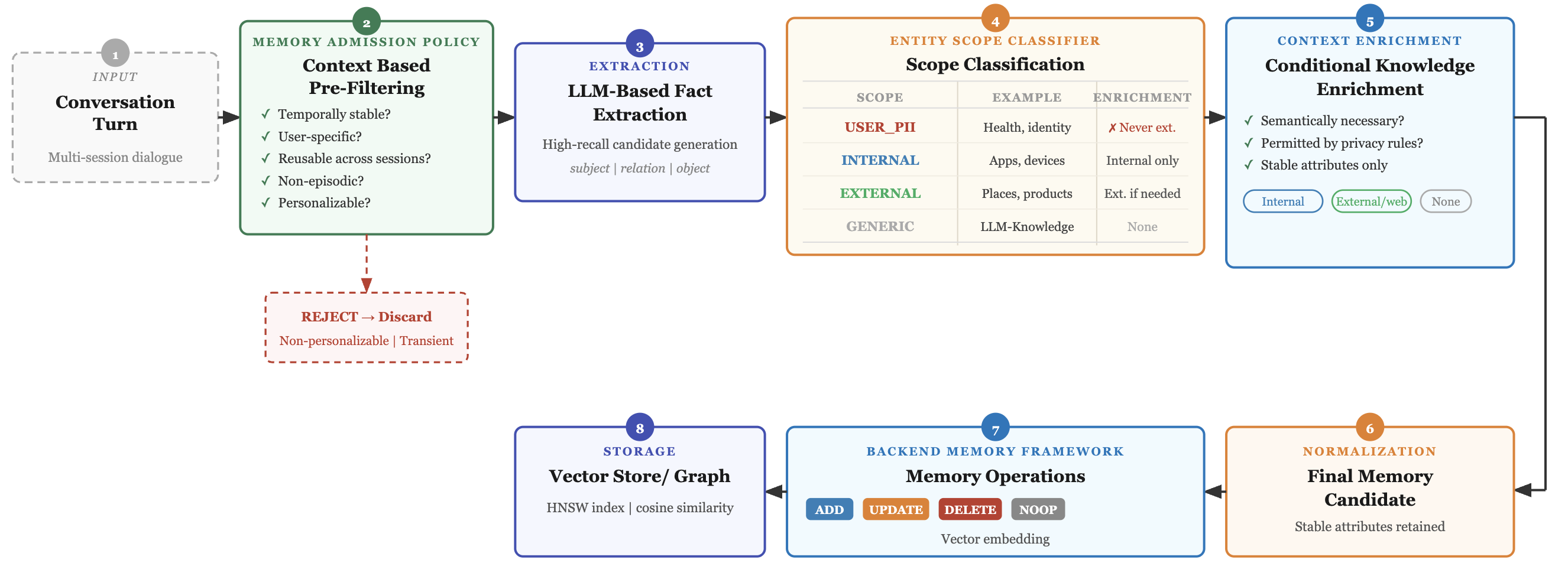}
  \caption{The Gated Memory pipeline. Every candidate fact passes through
  an utterance-level admission gate. Admitted facts are scope-classified and
  conditionally enriched before being passed to the memory backend.}
  \label{fig:pipeline}
\end{figure*}

Gated Memory is a drop-in formation layer that sits between conversation and
any existing memory backend, decomposing into three sequential stages:
admission, scope classification, and conditional enrichment
(Figure~\ref{fig:pipeline}). All three stages operate over a
\emph{current-versus-reference} view of the dialogue, described next.

\paragraph{Stage 0: Current--Reference Decomposition.}
A naive formation layer that consumes the entire visible dialogue history on
every turn re-evaluates, and therefore re-emits, facts it has already committed:
the same attribute is written on every subsequent turn in which it remains
visible, and unbound references in the history are resolved against the wrong
antecedent, fabricating entities that the current turn never mentioned. Gated
Memory eliminates this class of error by partitioning the input into a
\emph{current exchange} ,the single user utterance and its system response that
formation must evaluate,and a \emph{reference history} of prior turns that is
\emph{context-only}: it is available for pronoun and deixis resolution but is
never itself a source of new facts. Concretely, extraction is scoped to the
current exchange, while the reference history is supplied to the admission prompt
solely to bind references (\textit{``he,'' ``there,'' ``that one''}) to their
antecedents. This decomposition is what makes formation idempotent across turns:
a fact is formed once, when its evidence first appears, and is not re-formed
merely because it remains contextually visible. In practice the dialogue is
processed as overlapping rolling windows over the current exchange, with a small
turn overlap that supplies the local antecedents required for reference
resolution without expanding the set of turns treated as fact sources.

\paragraph{Stage 1: Utterance-Level Admission.}
Every candidate fact is evaluated by a structured LLM prompt against the
\emph{complete original utterance} ,not the extracted triple. This is the
load-bearing design choice that operates before extraction and retains
ownership context that no downstream component can recover.
A candidate is \textbf{admitted} if it satisfies at least one criterion:
(1) it encodes a personalization relation (preference, habit, ownership, role,
or stable affiliation),(2) it encodes a semi-stable user attribute (health, employment, location, or major life event),or (3) it exhibits
cross-session reusability ,it would materially improve relevance of a future
response. A candidate is \textbf{rejected} if it encodes a transient state, a
generic emotional reaction, or vague non-factual content. The following
contrasts illustrate why utterance-level evaluation is necessary ,each pair
produces identical extracted triples:

\begin{examplebox}
\small
\reject{``I drove  a Jeep for the road trip that I rented ;''}
\textit{transient,context lost at extraction}\\[3pt]
\admit{``I've had my Jeep Wrangler for three years.''}
\textit{stable ownership}\\[3pt]
\reject{``I might switch to Android next year.''}
\textit{hedged,no commitment}\\[3pt]
\admit{``I'm switching to Android next month.''}
\textit{definitive,stable preference}
\end{examplebox}

The admission policy outputs a structured JSON object containing the
admit/reject decision, entity based ontology mapping, scope classification, and enrichment flag. Every decision
is logged and auditable.

\paragraph{Multi-intent decomposition.}
A single admitted utterance frequently carries more than one storable
proposition, and committing it as one entangled string degrades both retrieval
precision and later conflict resolution. Formation therefore decomposes an
admitted exchange into one or more \emph{atomic} facts, each evaluated and
categorized independently. For example, \textit{``gift ideas for my son''}
yields both a persona fact (\textit{the user has a son}) and a procedural fact
(\textit{the user searched for gift ideas}); \textit{``find a hotel with an
onsen, and remember I'm vegetarian''} yields a scoped preference and a durable
persona attribute. Decomposition is performed at formation, while the utterance
is intact, because the boundaries between propositions,and the reference
resolution each requires, are themselves part of the context that extraction
would otherwise discard.

\paragraph{Formation categories.}
Each atomic fact is assigned one or more of six categories that determine how the
backend should treat it: \texttt{persona} (identity, relationships, durable
attributes), \texttt{preference} (likes, dislikes, dietary and interaction
constraints), \texttt{event} (experienced episodes, carrying temporal and
participant metadata), \texttt{procedural} (device- or application-level actions
the user performed), \texttt{user\_rule} (explicit instructions on assistant
behavior), and \texttt{error\_learning} (corrections to prior assistant
understanding). Categories are not mutually exclusive: a fact that aids retrieval
under more than one is multi-labeled rather than forced into a single class. The
categorization is produced jointly with admission so that it, too, is conditioned
on the full utterance rather than on the extracted triple.

\paragraph{Provenance.}
Every fact carries a \emph{provenance} label distinguishing a \emph{directly
stated} attribute, one the user asserts as the point of the utterance
(\textit{``I have a son''}), from an \emph{inferred} one, extracted from a
reference made in passing (\textit{``gift ideas for my son''} implies, but does
not assert, that the user has a son). This distinction, like ownership versus
rental, exists only in the original utterance and is unrecoverable downstream. It
governs how confidently the backend may treat a fact: directly stated attributes
are candidates for durable semantic storage, whereas inferred attributes are
retained provisionally, pending corroboration. Under explicit user instruction to
remember, all resulting facts are marked directly stated.

\paragraph{Grounding.}
Formation is constrained to \emph{assert only what the evaluated window
supports}. The policy may not introduce a specific entity or relation absent from
the current exchange and its reference context; an unbound reference is resolved
to the most specific accurate antecedent, or, failing that, to a generic form,
rather than to an invented one. For instance, a window containing only
\textit{``he is seven''} with no antecedent naming a child yields
\textit{the user's child is seven}, not a fabricated son. This grounding
constraint converts a silent failure mode of triple extraction,confident
assertion of unsupported entities,into a bounded, auditable decision. Formation
also preserves the \emph{language} of the source utterance, so that a fact stated
in one language is stored in that language rather than silently translated.

\paragraph{Stage 2: Entity Scope Classification.}
Every admitted fact is assigned to exactly one of four scopes
(Table~\ref{tab:scope}), produced jointly with the admission decision and privacy constraint classification. 

\begin{table}[t]
\centering
\scriptsize
\caption{Entity scope taxonomy and enrichment policy with privacy constraints.}
\label{tab:scope}
\setlength{\tabcolsep}{3pt}
\begin{tabular}{p{1.4cm} p{2.4cm} p{2.9cm}}
\toprule
\textbf{Scope} & \textbf{Covers} & \textbf{Enrichment policy} \\
\midrule
\texttt{USER\_PII} & Health, identity, finances, contacts &
  \textbf{No enrichment.} \\
\addlinespace[2pt]
\texttt{INTERNAL} & Devices, apps, platform data &
  Internal metadata only,no external calls \\
\addlinespace[2pt]
\texttt{EXTERNAL} & Products, places, public entities &
  External grounding when semantically necessary \\
\addlinespace[2pt]
\texttt{GENERIC} & LLM-resolvable facts &
  Parametric knowledge or none \\
\bottomrule
\end{tabular}
\end{table}

\paragraph{Stage 3: Conditional Enrichment.}
A fact is enriched only when two conditions are jointly satisfied:
enrichment would improve its semantic utility for future queries, and the
assigned scope permits the required knowledge source.
For \texttt{EXTERNAL} entities, a single stored fact can support a broad
class of downstream queries. For example:

\begin{examplebox}
\textit{``Planning a trip with BMW edrive''} $\rightarrow$ \\
\small\texttt{(Type:ElectricVehicle,}\\
\small\texttt{ChargingRange:``300 miles'')}
\end{examplebox}

\paragraph{Deictic grounding.}
Beyond entity-level enrichment, formation resolves relative temporal and spatial
references against the metadata of the current exchange. Expressions such as
\textit{``next week,'' ``tomorrow,''} or \textit{``tonight''} are grounded to
absolute dates using the utterance timestamp, and place deixis such as
\textit{``here''} or \textit{``there''} is resolved to a concrete location using
the request context, subject to the same scope-based privacy policy that governs
external enrichment. Grounding relative references at formation is, like entity
enrichment, a one-time cost amortized across every future retrieval: a fact whose
temporal anchor is resolved once remains interpretable indefinitely, whereas a
fact storing \textit{``next week''} verbatim becomes meaningless the moment the
utterance's temporal origin is lost.

After enrichment, facts are normalized. Volatile session-specific
context is stripped and only stable attributes are committed. Normalized,
enriched facts are passed to the memory backend, which handles deduplication, conflict
resolution, embedding, and persistence. Gated Memory governs what reaches the
backend, it does not replace it. Retrieval and generation are entirely
unchanged.

%% ────────────────────────────────────────────────────────────────
\section{Experiments}
\label{sec:experiments}

\subsection{Dataset}
We evaluate on LoCoMo-10, a benchmark of ten multi-session dialogues
averaging $\sim$300 turns each, with 1,540 questions across four categories:
Single-Hop, Multi-Hop, Temporal, and Open Domain. Category 5 (adversarial)
is excluded.

\subsection{Baseline and experimental design}
Our baseline is mem0 framework with 
a fixed configuration of Qdrant vector store, an open-source 8B 
instruction-tuned LLM and a fixed answer generation prompt. The only 
experimental variable is the presence of the Gated Memory 
formation layer, directly isolating formation quality from all 
other system components. Gated Memory is designed as a production formation layer. The baseline thus represents a complete 
ablation of the formation layer. Disentangling the individual 
contributions of admission and enrichment is left as future work.

\subsection{Metric}
We report \textbf{LLM-judge accuracy}. An independent evaluator scores each
answer as correct or incorrect against ground truth, with lenient matching
for paraphrasing and partial answers.

\subsection{Admission statistics}
The gate processed 5,882 utterances across ten conversations, admitting 5,527
(94.0\%) and rejecting 355 (6.0\%). Rejections consisted primarily of vague
emotional expressions, procedural turns, and transient situational references.

%% ────────────────────────────────────────────────────────────────
\section{Results and Discussion}
\label{sec:results}

\begin{table}[t]
\centering
\scriptsize
\caption{LLM-judge accuracy (\%) on LoCoMo-10. Retrieval, generation, and
evaluation are identical across both conditions.}
\label{tab:results}
\setlength{\tabcolsep}{4pt}
\begin{tabular}{lrrrrr}
\toprule
\textbf{System} & \textbf{S-Hop} & \textbf{M-Hop} & \textbf{O-Dom} & \textbf{Temp} & \textbf{Overall} \\
\midrule
memory baseline  & 49.6 & 50.0 & 48.3 & 57.6 & 50.6 \\
Gated Memory  & 51.8 & 47.9 & 49.9 & 58.6 & 51.9 \\
\midrule
Rel Error.\ $\Delta$ &
  \textcolor{admitgreen}{+4.4\%} &
  \textcolor{rejectred}{$-$4.2\%} &
  \textcolor{admitgreen}{+3.3\%} &
  \textcolor{admitgreen}{+1.7\%} &
  \textcolor{admitgreen}{+2.6\%} \\
\bottomrule
\end{tabular}
\end{table}

Table~\ref{tab:results} reports results. Gated Memory improves overall accuracy
by +2.6\% relative with the formation layer as the sole experimental variable,
directly attributing the gain to formation quality.

\paragraph{Single-Hop and Open Domain.}
The largest gains appear on Single-Hop (+4.4\%) and Open Domain (+3.3\%).
Single-Hop performance is directly sensitive to retrieval noise: removing
low-signal entries reduces competition for top-$k$ slots, improving
precision over the matched retrieval pool. Open Domain additionally benefits
from conditional enrichment ,an EV entry augmented with enriched context supports a broad class of queries that the raw
extraction string cannot address, and the +3.3\% gain quantifies this effect.

\paragraph{Temporal.}
The +1.7\% gain on Temporal questions is consistent with the admission
policy's preference for temporally stable facts over episodic states. Facts
encoding persistent preferences, and semantic relevance are admitted,transient situational references are rejected.

\paragraph{Multi-Hop regression.}
Multi-Hop exhibits a $-$4.2\% regression. Compositional reasoning is sensitive
to both precision \emph{and} recall. The admission
policy, optimized for individual fact utility, occasionally prunes interdependent nodes.
This is a precision-recall tradeoff inherent to independent fact evaluation,
not an artifact of our specific threshold, and motivates \emph{chain-aware
admission} that models inter-fact dependencies at formation time as the primary
direction for future work.

\paragraph{Formation quality as a binding constraint.}
With retrieval, generation, and evaluation held constant, a +2.6\% overall
gain from changing only what enters the memory store especially for a emotionally dense dataset demonstrates that
\emph{formation quality is a binding constraint on downstream task performance}. In actual production environments with high volume of irrelevant context this gating framework could be much more beneficial.

%% ────────────────────────────────────────────────────────────────
\section{Related Work}
\label{sec:related}
Persistent memory is a very critical component for AI systems ~\citet{timoneda2025memory, he2024human}.
Long-term memory architectures for conversational AI have focused mostly on
post-formation concerns ~\citet{zhong2024memorybank, liu2023think}. MemGPT~\citet{memgpt} introduces hierarchical memory
management with OS-inspired paging while mem0~\citet{mem0} provides production
lifecycle management including deduplication and conflict resolution.
A-MEM~\citet{amem} proposes Zettelkasten-inspired dynamic interconnection while
ReadAgent~\citet{readagent} addresses long-context limits through gist-based
compression. Zep ~\citet{rasmussen2025zep} uses Graphiti, a temporally-aware knowledge graph engine to dynamically synthesizes both unstructured and structured data while maintaining historical relationships. None of this frameworks addresses the formation quality. 
Prior Works on retrieval-augmented generation~\citet{lewis2020rag} and knowledge
conflict~\citet{shi2023distracted} has also established that retrieval noise
degrades generation quality. Our results extend this causal chain: formation
noise $\rightarrow$ retrieval noise $\rightarrow$ generation degradation.
Classic work on memory decay~\citet{tulving1985,baddeley2000} addresses
\emph{when} facts should be removed,we address \emph{whether} they should even be
admitted. The episodic--semantic distinction~\citet{tulving1985} is typically
applied \emph{post hoc}, as a categorization of already-stored content; we instead
determine category, provenance, and applicability scope \emph{at formation}, while
the utterance is intact, because the signals that separate a directly stated
attribute from one inferred in passing, or an episodic event from a durable
persona attribute, are properties of the original utterance rather than of the
extracted triple. Prior lifecycle-oriented systems~\citet{memgpt,mem0} operate on
extracted triples and cannot recover these signals; the grounding constraint we
impose,asserting only entities present in the evaluated context,further addresses
a fabrication failure mode of triple extraction that downstream deduplication,
which only detects conflicts and duplicates, cannot catch. Privacy research at the training and inference
level~\citet{carlini2021} motivates but does not address risks introduced by
enrichment,our entity scope taxonomy and its policy based enrichment fills
this gap.

%% ────────────────────────────────────────────────────────────────
\section{Conclusion}
\label{sec:conclusion}
We introduced Gated Memory, a formation framework that addresses an unexamined
failure point in long-term conversational memory. By placing an utterance-level admission gate at the moment a fact is first
written to storage and a
privacy-aware enrichment stage before extraction, this framework preserves contextual signals
irreversibly lost under existing approaches for better memory formation. A +2.6\% relative improvement on
LoCoMo-10 with no changes to retrieval, generation, or evaluation establishes
formation quality as a measurable constraint on memory performance, though a
$-$4.2\% regression on Multi-Hop reasoning reveals a precision-recall tradeoff
that chain-aware, inter-fact-dependent admission policies should resolve. \textbf{In production settings, the gate's filtering of vacuous 
content directly reduces storage costs, retrieval latency, 
and hallucination risk at scale}. Domain-adaptive thresholds 
and knowledge-graph grounding remain natural extensions of future work. The richness of what a personalized assistant
can do is bounded by the quality of what it chooses to remember.

\section*{Limitations}
The admission gate's effectiveness depends on the quality of 
the underlying LLM prompt. Adversarial or ambiguous utterances 
may be misclassified at formation time. Unlike downstream 
deduplication, formation errors are undetectable after the fact is formed. 
The evaluation is conducted on LoCoMo-10 with atypical emotional density in utterance data. Hence generalization 
across domains, languages, and task-oriented deployments need 
to be demonstrated empirically. The 6.0\% rejection rate is a 
conservative lower bound and task-oriented deployments are expected 
to exhibit substantially higher filtering rates that needs to be analyzed further. The Multi-Hop regression ($-$4.2\%) 
reveals a precision-recall tradeoff inherent to independent fact 
evaluation,chain-aware admission that models inter-fact 
dependencies remains future work. Finally, the enrichment stage 
introduces external knowledge calls for \texttt{EXTERNAL}-scoped 
facts, adding latency and potential staleness risk for 
time-sensitive information that needs to be addressed by only including stable facts for enrichment.

%% ────────────────────────────────────────────────────────────────
\bibliography{refs}

\end{document}